\documentclass[11pt,a4paper]{article}
\usepackage[hyperref]{acl2020}
\usepackage{times}
\usepackage{latexsym}
\usepackage{makecell}
\usepackage{graphicx}
\usepackage{comment}

\graphicspath{ {./figure/} }

\usepackage{microtype}

\aclfinalcopy 

\DeclareRobustCommand\onedot{\futurelet\@let@token\@onedot}
\def\onedot{. } 
\def\eg{\emph{e.g}\onedot} 
\def\ie{\emph{i.e}\onedot}

\title{Augmenting Data for Sarcasm Detection \\with Unlabeled Conversation Context}

\author{Hankyol Lee \\
  RippleAI\\
  Seoul, Korea\\
  {\tt\small hklee@rippleai.co} \\\And
  Youngjae Yu \\
  RippleAI \& SNU\\
  Seoul, Korea\\
  {\tt\small yj.yu@rippleai.co} \\\And
  Gunhee Kim \\
  RippleAI \& SNU \\
  Seoul, Korea\\
  {\tt\small gunhee@snu.ac.kr}
}

\date{}

\begin{document}

\maketitle
\begin{abstract}
  We present a novel data augmentation technique, CRA (Contextual Response Augmentation), which utilizes conversational context to generate meaningful samples for training. We also mitigate the issues regarding unbalanced context lengths by changing the input-output format of the model such that it can deal with varying context lengths effectively.
Specifically, our proposed model, trained with the proposed data augmentation technique, 
participated in the sarcasm detection task of FigLang2020, have won and achieves the best performance in both Reddit and Twitter datasets.
\end{abstract}

\section{Introduction}
The performance of many NLP systems largely depends on their ability to understand figurative languages such as irony, sarcasm, and metaphor \citep{pozzi2016sentiment}. 
The results from the Sentiment Analysis task held in SemEval-2014 \citep{martinez2014sentiment}, for example, show that apparent performance drops occur when the figurative language is involved in the task.
This work aims, in particular, to design a model that identifies sarcasm in the conversational context. 
More specifically, the goal is to determine whether a response is sarcastic or not, given the immediate context (\ie only the previous dialogue turn) and/or the full dialogue thread (if available).
For evaluation of our model, we participated in the FigLang2020 sarcasm challenge\footnote{\url{https://competitions.codalab.org/competitions/22247}.}, and have won the competition as our model is ranked 1 out of 35 teams for the Twitter dataset and 1 out of 34 teams for the Reddit dataset.

We summarize our technical contributions to win the challenge as follows:
\begin{enumerate}
\item We propose a new data augmentation technique that can successfully leverage the structural patterns of the conversational dataset. Our technique, called CRA(Contextual Response Augmentation), utilizes the conversational context of the unlabeled dataset to generate new training samples. 
\item  The context lengths  (\ie previous dialogue turns) are highly variable across the dataset.
To cope with such imbalance, we propose a context ensemble method that exploits multiple context lengths to train the model. The proposed format is easily applicable to any Transformer \citep{vaswani2017attention} encoders without changing any model architecture.
\item We propose an architecture where the Transformer Encoder is stacked with BiLSTM~\citep{schuster1997bidirectional} and NeXtVLAD~\citep{lin2018nextvlad}. We observe that NeXtVLAD, a differentiable pooling layer, proves more effective than simple nonparametric mean/max pooling methods.
\end{enumerate}

\section{Approach}

\begin{figure}[t]
   \centering
   \includegraphics[width=0.43\textwidth]{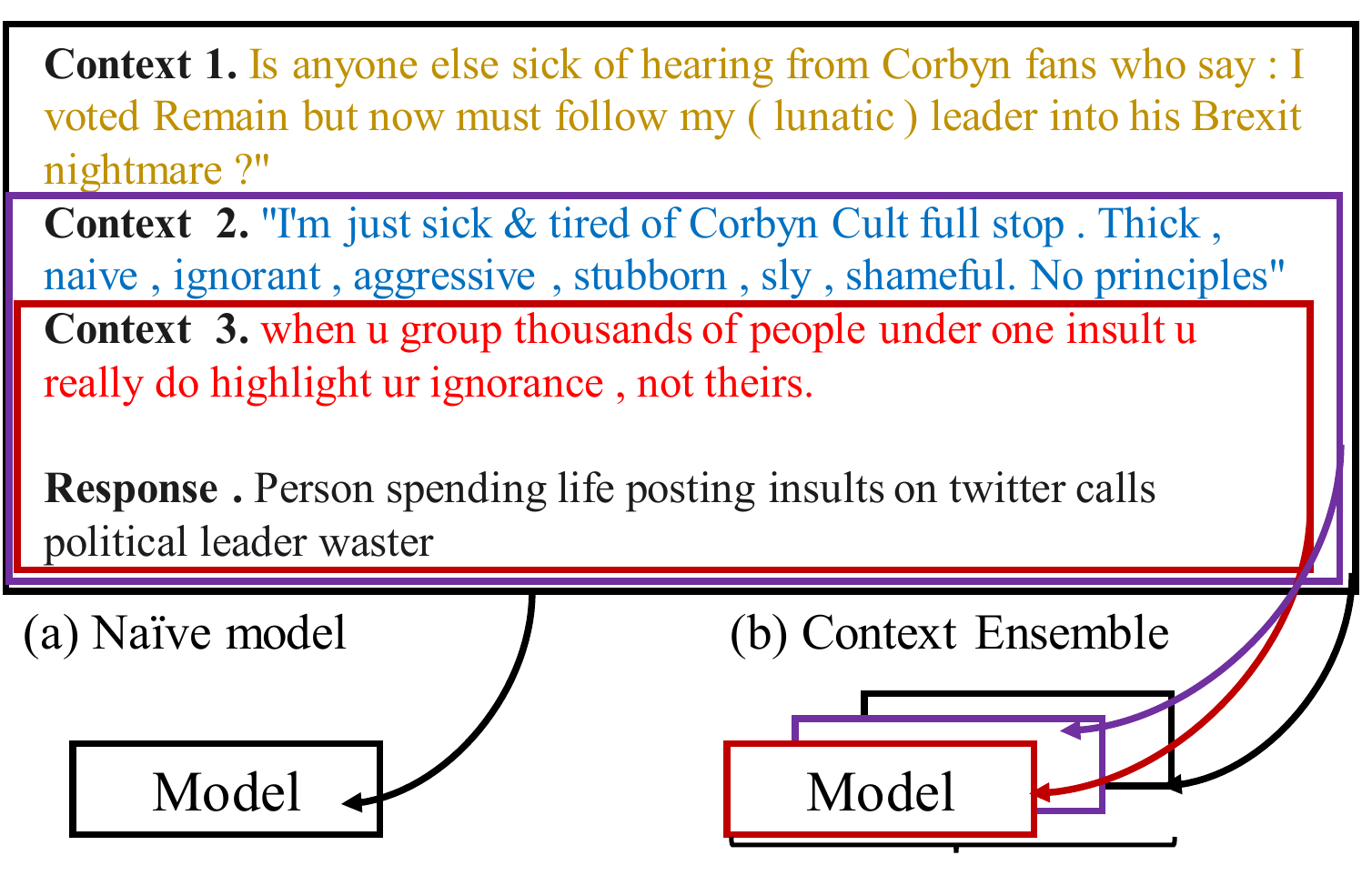} 
   \caption{Illustration of the context ensemble method for Sarcasm detection. We train multiple models with different context window sizes, and ensemble them for inference.}
   \label{fig:ensemble}
   \vspace{-5pt}
\end{figure}

The task of our interest is, given response ($r_1$) and its previous conversational context ($c_{1}, c_{2}, \cdots, c_{n}$),
to predict whether the response $r_1$ is sarcastic or not (See an example in  Figure \ref{fig:ensemble}). 
We below discuss our model (section \ref{sec:model}), training details (section \ref{sec:training})
and the proposed data augmentation techniques (section \ref{sec:augmentation}).

\begin{figure}[t]
   \centering
   \includegraphics[width=0.43\textwidth]{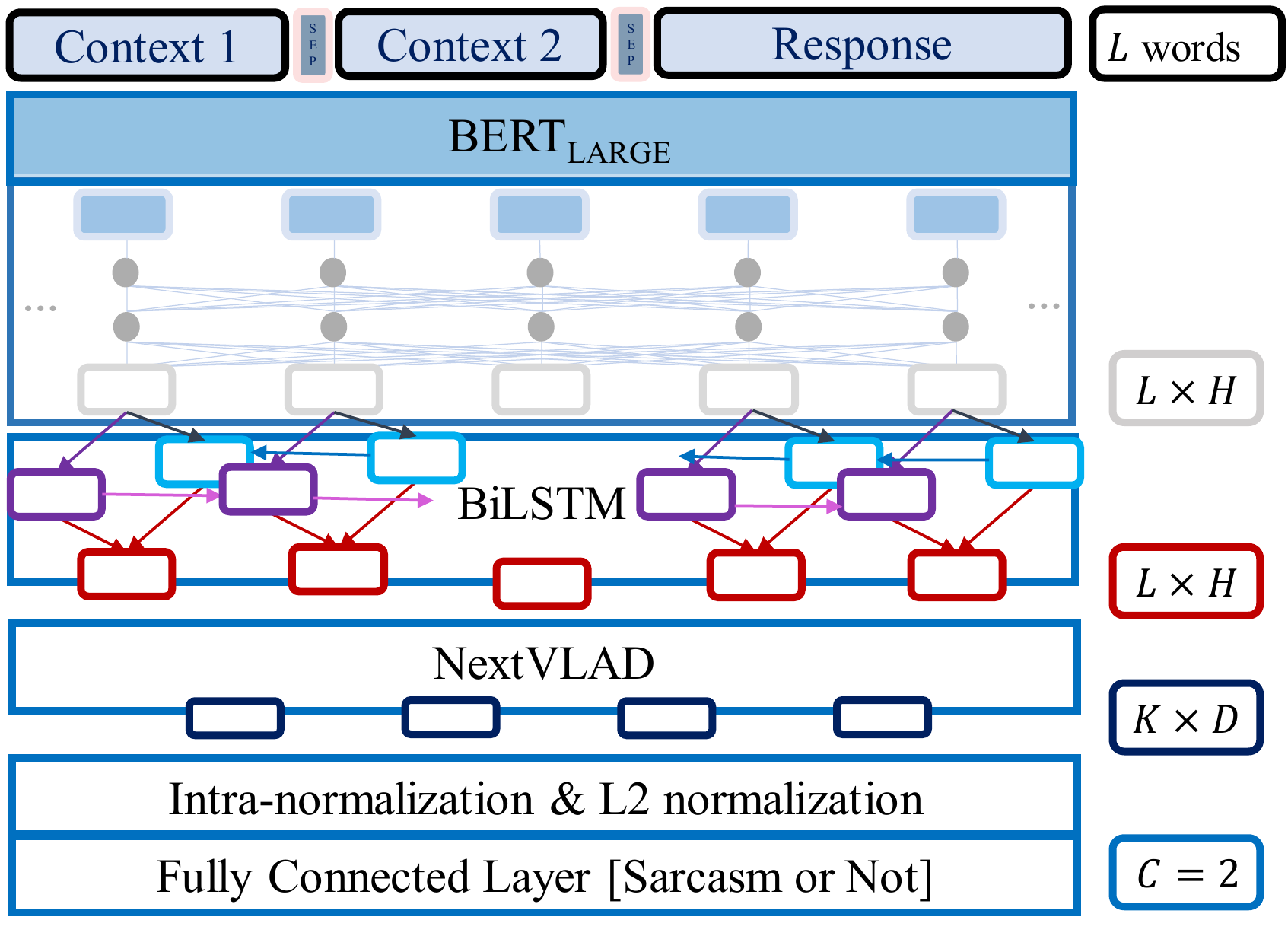} 
   \caption{The architecture of our best performing model for sarcasm detection.}
   \label{fig:arch}
   \vspace{-5pt}
\end{figure}

\begin{table}[b]
\small
\centering
\setlength\tabcolsep{5.5pt}
\vspace{-5pt}
\begin{tabular}{l|c|c|c}
\hline
 & $train_{data}$ & $valid_{data}$ & $test_{data}$ \\ \hline
Twitter & 4000 & 1000 & 1800 \\
Reddit  & 3520 & 880 & 1800 \\
\hline
\end{tabular}
\vspace{-5pt}
\caption{Dataset Splitting.}
\label{tab:data_division_detail}
\vspace{-8pt}
\end{table}

\subsection{The Model}
\label{sec:model}

Figure~\ref{fig:arch} describes the architecture of our best performing model.
The model broadly consists of two parts: the transformer (BERT) \citep{devlin2018bert} and pooling layers, which are decomposed into BiLSTM \citep{schuster1997bidirectional} and NetXtVLAD \citep{lin2018nextvlad} as an improved version of NetVLAD \citep{arandjelovic2016netvlad}. Reportedly, NetVLAD is a CNN-model that is highly effective and more resistant to over-fitting than usual temporal models such as LSTM or GRU \citep{lin2018nextvlad}.
The Implementation of these models are as follows:
\begin{itemize}
\item BERT(large-cased): 24-layer, 1024-hidden and 16-heads.
\item BiLSTM: 2-layer, 1024-hidden and 0.25-dropout.
\item NeXtVLAD: 8-groups, 4-expansion, 128-number of clusters and 512-cluster size.
\end{itemize}

\subsection{Training Details}
\label{sec:training}

We use the entropy loss on the last softmax layer in the model. The training batch size is 4 for all the experiments. We adopt the cyclic learning rate \citep{smith2017cyclical}, where the initial learning rate is 1e-6, and the moment parameters are (0.825, 0.725).

\textbf{Dataset Splitting}.
We further split the provided training set ($training_{data}$) into the training ($train_{data}$) and validation ($valid_{data}$) set as in Table~\ref{tab:data_division_detail}. We use $valid_{data}$ for early stopping and the model performance validation during the training phase.

\textbf{Context Ensemble}.
Figure~\ref{fig:ensemble} depicts the idea of the context ensemble method to cope with highly variable context lengths in the dataset.
Instead of using the training data as their original forms only (Figure~\ref{fig:ensemble}(a)),
we consider multiple context window sizes as separate data, which can naturally balance out the proportion of short and long context (Figure~\ref{fig:ensemble}(b)).

\begin{figure}[t]
   \centering
   \includegraphics[width=0.45\textwidth]{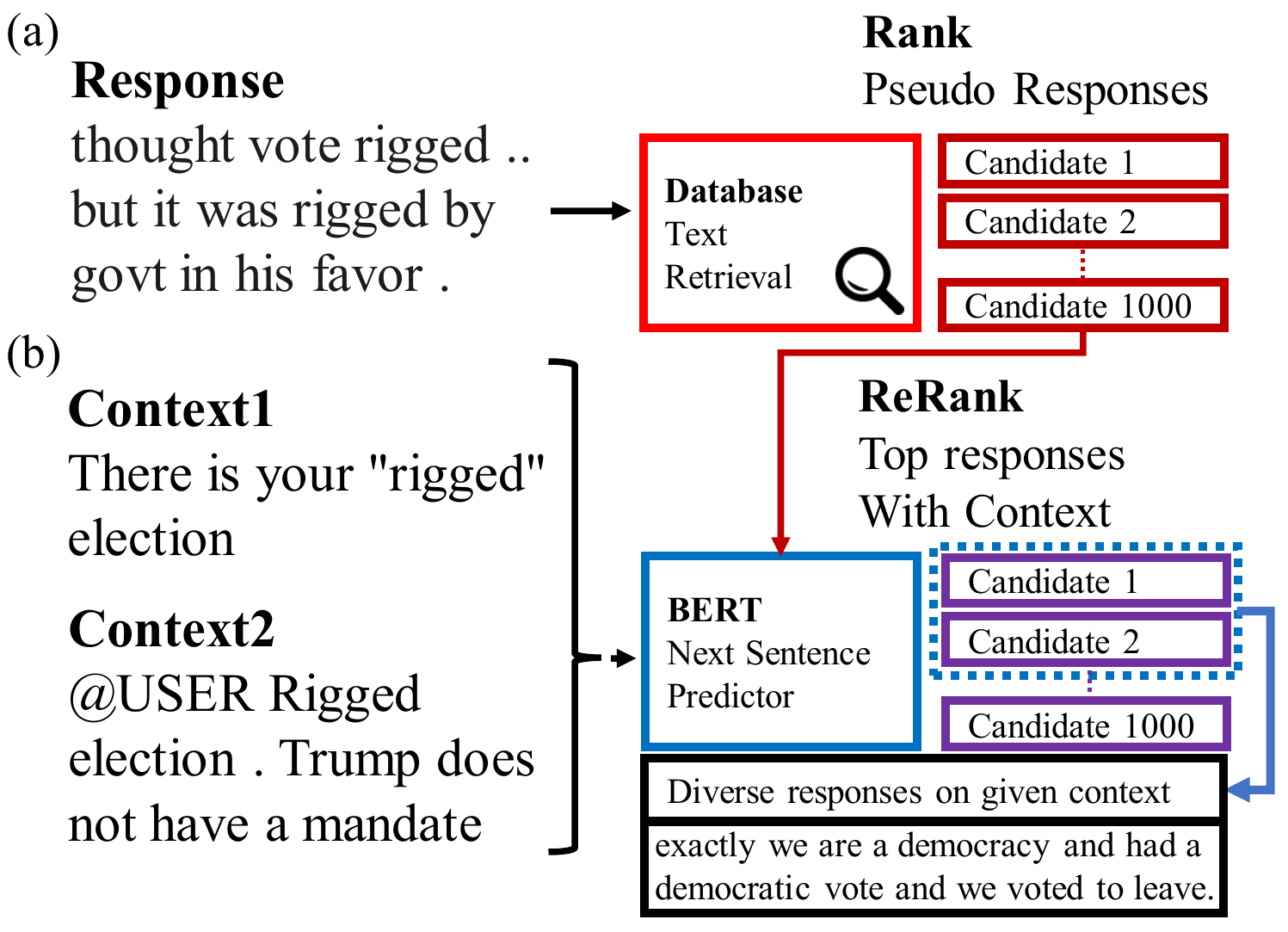} 
   \caption{Overview of the proposed Contextual Response Augmentation (CRA). Using (a) Text query retrieval on sarcasm database and (b) Reranking best responses conditioned on a given context, we obtain various pseudo responses that are useful for training.}
   \label{fig:cra}
   \vspace{-5pt}
 \end{figure}

\begin{table*}[t!]
\small
\centering
\setlength\tabcolsep{5.5pt}
\begin{tabular}{l|l}
\hline
\multicolumn{2}{c}{An unsarcastic sample} \\ \hline
c1 & Dont mind me, its just a gun \\ 
c2 & The dude in the front row is like 'Are we gonna do something?' \\ 
c3 & \makecell[l]{It's the perfect example of how the bystander effect works, even amongst police (or whatever they are) \\ 
     100\% if they were alone and saw this they'd doay something to the guy. \\
     But together you get a pack mentality.}\\ \hline
r1 & Unfortunately, the police are sometimes the victimizers   \\ \hline
\multicolumn{2}{c}{A sarcastic sample} \\ \hline
 c1 & \makecell[l]{Trump won Wisconsin by 27,000 votes. 300,000 voters were turned away by the states strict Voter ID law. \\ There is your \"rigged\" election ."} \\ 
c2 & @USER Rigged election . Trump does not have a mandate . Period. \\ \hline
r1 & @USER @USER @USER exactly we are a democracy and had a democratic vote and we voted to leave \\
\hline
\end{tabular}
\caption{Samples generated from unlabeled dataset}
\vspace{-5pt}
\label{tab:Generated_samples}
\end{table*}
\begin{table*}[t]
\small
\centering
\setlength\tabcolsep{5.5pt}
\begin{tabular}{l|cc|cc|cc}
\hline
Metric &
\multicolumn{2}{c|}{Precision} &
\multicolumn{2}{c|}{Recall} &
\multicolumn{2}{c}{F1} \\
\hline
dataset  & $twitter_{valid}$ & $reddit_{valid}$ & $twitter_{valid}$ & $reddit_{valid}$ & $twitter_{valid}$ & $reddit_{valid}$ \\ \hline
T+BiLSTM+NeXtVLAD & 0.8295 & 0.6414 & 0.8816 & 0.7867 & 0.8548 & 0.7067 \\
T+BiLSTM+MaxPool  & 0.8558 & 0.6620 & 0.8182 & 0.7092 & 0.8366 & 0.6848 \\ 
T+BiLSTM+MeanPool & 0.7339 & 0.6881 & 0.8683 & 0.5837 & 0.7954 & 0.6316 \\
T+NeXtVLAD        & 0.8163 & 0.5891 & 0.8785 & 0.7976 & 0.8462 & 0.6777 \\ 
\hline \hline
T+BiLSTM+NeXtVLAD & 0.8747 & 0.6938 & 0.9219 & 0.8187 & 0.8977 & 0.7513 \\
T+BiLSTM+MaxPool  & 0.8318 & 0.6624 & 0.8751 & 0.7910 & 0.8529 & 0.7210 \\ 
T+BiLSTM+MeanPool & 0.7856 & 0.6089 & 0.8792 & 0.8070 & 0.8298 & 0.6941  \\
T+NeXtVLAD        & 0.8525 & 0.6888 & 0.9101 & 0.7792 & 0.8804 & 0.7313  \\ 
\hline
\end{tabular}
\caption{Sarcasm detection performance on the validation set.
 The upper and lower part of the table respectively denote the performance before and after data augmentation is applied. We set the context length to 3 for all models.}
\label{tab:exp_result}
\vspace{-5pt}
\end{table*}

\subsection{Data Augmentation}
\label{sec:augmentation}

\citet{van2018semeval} and \citet{ilic2018deep} have observed that in the case of Twitter, fueling additional data from the same domain did not help much the performance for detecting sarcasm and irony. However, this does not mean that the data augmentation would fail to improve sarcasm detection.
We use two techniques to augment the training data according to whether the data are labeled or not.
Especially, our data augmentation method named Contextual Response Augmentation (CRA) can take advantage of unlabeled dialogue threads, which are abundant and cheaply collectible. Figure \ref{fig:cra} illustrates the overview of our CRA method whose details are presented in section \ref{sec:aug_unlabeled}.

\subsubsection{Augmentation with Labeled Data}
Each training sample consists of contextual utterances, a response and its label ("SARCASM" or "NOT\_SARCASM"): $[c_{1}, c_{2}, \cdots, c_{n}, r_{1}, l_1]$.
Our idea is to take the context sequence $[c_{1}, c_{2}, \cdots, c_{n}]$ as a new datapoint and label it as "NOT\_SARCASM". 
As shown in Figure~\ref{fig:ensemble}, without the response $[r_1]$, the sequence could not be labeled as "SARCASM". We hypothesize that these newly generated negative samples help the model better focus on the relationship between the response $[r_1]$ and its contexts $[c_{1}, c_{2}, \cdots, c_{n}]$. Also, we balance out the number of negative samples by creating positive samples via back-translation methods (\citet{berard2019naver}; \citet{zheng2019robust}), which simply translate the sentences into another language and then back to the original language to obtain possibly rephrased data points. For the back-translation, we have used 3 languages [French, Spanish, Dutch].

\subsubsection{Augmentation with Unlabeled Data}
\label{sec:aug_unlabeled}
We also generate additional training samples using the unlabeled data: $[c_{1}, c_{2}, \cdots, c_{n}, r_{1}]$.
This approach is tremendously useful since a huge amount of unlabeled dialogue threads can be collected at little cost.
As shown in Figure \ref{fig:cra}, the procedures for unlabeled augmentation are as follows:

\begin{enumerate}
    \item We encode each response in the labeled training set using the BERT trained on natural inference tasks \citep{reimers-2019-sentence-bert}.
    \item Given unlabeled data $[c_{1}, c_{2}, \cdots, c_{n}, r_{1}]$, we encode $[r_{1}]$ and find the most similar top $k (=1000)$ data from the labeled database. We denote them as $\{r_{t,1}, \cdots, r_{t,k}\}$. 
    \item We rank the top $k$ candidates according to the next sentence prediction (NSP) confidence of BERT\footnote{We fine-tune BERT only for the next sentence prediction task using the corpora in Table \ref{tab:dataset_sizes} and the $training_{data}$}. That is, we input $[c_{1}, c_{2}, \cdots, c_{n}, sep, r_{t,i}]$ to BERT, and compute the NSP confidence of $r_{t,i}$  for all $i \in \{1,\cdots, k\}$. We then select the most confident response $r_t^*$ with its label $l_t^*$ and make a new data point $[c_{1}, c_{2}, \cdots, c_{n}, r_{t}^*, l_t^*]$. 
\end{enumerate}

Table~\ref{tab:Generated_samples} shows some samples generated from this technique. The quality of generated data depends undoubtedly on the degree of contextual conformity and similarity between the initial responses.
We find, however, that adding more data makes the quality of the augmented data better as the label transfer noise becomes attenuated. In summary, besides the standard datasets shown in Table~\ref{tab:dataset_sizes}, we further crawled 100,000 texts from both Twitter and Reddit for the augmentation with unlabeled data.

\begin{table}[t]
\small
\centering
\setlength\tabcolsep{5.5pt}
\begin{tabular}{l|ccc}
\hline
Teams & Precision & Recall & F1 \\
\hline
\textbf{miroblog} & \textbf{0.932} & \textbf{0.936} & \textbf{0.931} \\
nclabj  & 0.792 & 0.793 & 0.791 \\ 
Andy3223 & 0.7910 & 0.7940 & 0.790 \\
DeepBlueAI        & 0.78 & 0.785 & 0.779 \\ 
ad6398        & 0.773 & 0.775 & 0.772 \\
\hline \hline
\textbf{miroblog} & \textbf{0.834} & \textbf{0.838} & \textbf{0.834}  \\
Andy3223  & 0.751 & 0.755 & 0.75  \\ 
DeepBlueAI & 0.749 & 0.750 & 0.749   \\
kevintest        & 0.746 & 0.746 & 0.746  \\
Taha        & 0.738 & 0.739 & 0.737  \\
\hline
\end{tabular}
\caption{The FigLang2020 Sarcasm Scoreboard for Twitter (upper) and Reddit (below) dataset.
Our method miroblog achieves the best performance in both datasets with significant margins.}
\label{tab:scoreboard}
\vspace{-5pt}
\end{table}

\begin{table}[t]
\small
\centering
\setlength\tabcolsep{5.5pt}
\begin{tabular}{l|c|c|c}
\hline
$twitter_{valid}$ & Precision & Recall & F1 \\ \hline
no augmentation & 0.8294 & 0.8816 & 0.8547 \\
labeled augmentation & 0.8676 & 0.8550 & 0.8613 \\
unlabeled augmentation & 0.8747 & 0.9219 & 0.8977 \\
\hline
\end{tabular}
\caption{Sarcasm detection performance according to data augmentation on the $twitter_{valid}$ dataset.}
\label{tab:exp_aug}
\end{table}
\begin{table}[t]
\small
\centering
\setlength\tabcolsep{5.5pt}
\begin{tabular}{l|c|c|c}
\hline
$twitter_{valid}$ & Precision & Recall & F1 \\ \hline
Ensemble (max context) & 0.8558 & 0.8182 & 0.8366 \\
Ensemble (3 context)   & 0.8320 & 0.8288 & 0.8304 \\
Single (3 context)     & 0.8147 & 0.8052 & 0.8099 \\
\hline
\end{tabular}
\caption{Sarcasm detection performance according to the ensemble methods on the $twitter_{valid}$ dataset.}
\label{tab:exp_ensemble}
\vspace{-5pt}
\end{table}

\section{Experiments}

We first report the quantitative results by referring to the statistics in the official evaluation server \footnote{\url{https://competitions.codalab.org/competitions/22247}.} of the FigLang2020 sarcasm challenge as of the challenge deadline (\ie April 16, 2020, 11:59 p.m. UTC).
Table \ref{tab:scoreboard} summarizes the results of the competition, where our method named miroblog shows significantly better performance than other participants in both Twitter and Reddit dataset.
We report Precision (P), Recall (R), and F1 scores as the official metrics.

\begin{table}[t]
\small
\centering
\setlength\tabcolsep{5.5pt}
\begin{tabular}{c|c|c}
\hline
Reference         & Name              & Size    \\ \hline
\citet{ptavcek2014sarcasm}  & Platek             & 57041   \\ 
\citet{riloff2013sarcasm} & Riloff             & 1570    \\ 
\citet{khodak2017large}   & SARC-v2            & 321748  \\ 
\citet{khodak2017large}   & SARC-v2-pol        & 14340   \\ 
\citet{van2018semeval} & SemEval-2018-irony & 3851    \\ 
-                   & Web Crawled        & 100000  \\
\hline
\end{tabular}
\caption{The standard datasets and the crawled dataset (for unlabeled augmentation) used in the experiments.}
\label{tab:dataset_sizes}
\vspace{-5pt}
\end{table}

\subsection{Further Analysis}

We perform further empirical analysis to demonstrate the effectiveness of the proposed ideas.
We compare different configurations of pooling layers, context ensemble, and data augmentation.

\textbf{Pooling Layers}.
Table~\ref{tab:exp_result} shows the comparison of sarcasm detection performance between NeXtVLAD and other pooling methods in performance.
When coupled with BiLSTM, NeXtVLAD achieves better performance than max, and mean pooling methods.  

\textbf{Context Ensemble}.
Table~\ref{tab:exp_ensemble} shows the comparison with different context ensemble methods.
We use the baseline (Transformer+BiLSTM+ Maxpooling) and train it without augmenting the training set.
F1 scores of the model are better in the order of (a) ensemble with maximum context, (b) ensemble with three contexts and (c) no context.
The performance gap with or without context ensemble implies that balancing out the samples in terms of context length is important.
On the other hand, the performance gap between (a) and (b) is only 0.006, indicating that the use of older than three recent conversational contexts is scarcely helpful.

\textbf{Data Augmentation}.
Table~\ref{tab:exp_aug} compares the sarcasm detection results when the data augmentation is applied or not.
The augmentation with labeled data increases the F1 score from 0.854 to 0.861.
The augmentation with unlabeled data further enhances performance from 0.861 to 0.897.
The results demonstrate that both augmentation techniques help with the performance.

\begin{table*}[t!]
\small
\centering
\setlength\tabcolsep{5.5pt}
\begin{tabular}{l|l}
\hline
\multicolumn{2}{c}{\textbf{(i) The prediction is wrong without DA but correct with DA.}} \\ \hline
c1 & Any practice could be anyone's last practice. Yes. \\ 
c2 & @USER report: tom brady struck by lighting after leaving practice. \\ \hline
r1 & \textbf{[SARCASM]} @USER Report: Tom Brady abducted by space aliens during practice. \#NotReally \#Relax   \\ \hline
\multicolumn{2}{c}{\textbf{(ii) The prediction is correct without DA but wrong with DA}} \\ \hline
 c1 & \makecell[l]{@USER @USER @USER The racist trump is a Russian puppet.\\ He's a loser who's trying to destroy our constitution and hand this Country over to Putin.\\ He steals with the help of his white nationalist supporters. \\ He should be removed from Office and put in prison.} \\ 
c2 & \makecell[l]{@USER @USER @USER And who's drinking the koolaide ?\\ Mueller said no collusion or obstruction after spending \$ 30 million investigating - \\ with full access to the White House.\\ White Nationists \ unsubstantiated conspiracy theory.\\ Trump will win 2020 because people see him succeed through the nonsense.} \\ \hline
r1 & \makecell[l]{\textbf{[NOT\_SARCASM]} @USER @USER @USER You didn't bother to read the Mueller report, did you?\\ It was Barr who falsely exonerated your beloved cult leader. Read the Mueller report.\\ Until you do, don't propagate this lie. Educate yourself and read the report or shut up.\\ You ’ ll believe anything except the truth.} \\
\hline
r2 & \makecell[l]{\textbf{[SARCASM]} you not worry i are so blind, deaf. \\ I KNOW you have lost your sight (with regard)  \\listened to your cult leaders and Faux News and some Republicans.} \\
\hline
\multicolumn{2}{c}{\textbf{(iii) The prediction is wrong with and without DA.}} \\ \hline
c1 & \makecell[l]{I love this land called America \#VPDExperiment \#VPDDay \\ @USER and @USER at @USER. \\ The 30 Best Things to do in Washington DC: URL} \\ 
c2 & \makecell[l]{@USER @USER @USER Makes me just want to bow out of this whole thing right now... LOL} \\ 
c3 & \makecell[l]{@USER @USER @USER Noooooo! It's just the way I edit.\\ I'm trying all sorts of styles this 30 days.\\ No competition being done.}\\ \hline
r1 & \makecell[l]{\textbf{[SARCASM]} @USER @USER @USER Sorry, I forgot to use the font!\\ I'm loving your videos. Its giving me ideas and inspiration for some stuff I'd like to try.}   \\ \hline
\end{tabular}
\caption{Examples of three cases where data augmentation helps, hurts, or fails to improve the sarcasm predction.}
\vspace{-5pt}
\label{tab:Error analysis}
\end{table*}

\subsection{Error Analysis}
In order to better understand when our data augmentation methods are effective, we further analyze some examples of the following three cases according to whether the proposed labeled and unlabeled data augmentation (DA) is applied or not: (i) the prediction is wrong without DA but correct with DA, (ii) the prediction is correct without DA but wrong with DA, and (iii) the prediction is wrong with and without DA.
In other words, (i) is the case where DA helps, (ii) is the one where DA hurts, and (iii) is the one where DA fails to improve. 
 
Table~\ref{tab:Error analysis} shows some examples of these three cases. (i) The initial steps of the CRA involve finding similar training samples from the labeled database. Thus, after applying CRA, samples containing specific hashtags, \eg \#NotReally \#Relax, are included in the training set. We observe that theses tags tend to occur with the samples that are labeled ``SARCASM'', and thus CRA helps the model learn the correlation between the hashtags and the labels.
(ii) The augmented response ($r2$) contains the phrase ``cult leader'' as in the original response ($r1$). The corresponding label, however, is “SARCASM”. When the newly added samples do not match the context, or the labels are incorrect, CRA degrades the prediction.
(iii) The third case arises mostly when the situation is subtle and requires external knowledge beyond the given context. In order for the model to correctly classify the response as ``SARCASM'', the model requires to understand the tag \#VPD(Video Per Day). It is not clear what \#VPD is from the context, and without such knowledge, the model may still make incorrect predictions. 

\section{Conclusion}
We proposed a new data augmentation technique, CRA (Contextual Response Augmentation), that utilizes the conversational context of the unlabeled data to generate meaningful training samples.
We demonstrated that the method boosts the performance of sarcasm detection significantly.
The employment of both augmentations with labeled and unlabeled data enables the system to achieve the best F1 scores to win the FigLang2020 sarcasm challenge on both datasets of Twitter and Reddit. 

\bibliography{anthology,acl2020}
\bibliographystyle{acl_natbib}
\end{document}